\PassOptionsToPackage{hyphens}{url}
\documentclass[11pt,a4paper]{article}
\usepackage[hyperref]{ranlp2023}
\usepackage{times}
\usepackage{latexsym}

\usepackage{enumitem}
\setlist{noitemsep,nolistsep}
\usepackage{graphicx}
\usepackage{multirow}
\usepackage{bm}

\usepackage{microtype}

\aclfinalcopy % Uncomment this line for the final submission
\title{Mitigating Bias in Text Classification via Prompt-Based Text Transformation}

\author{Charmaine Barker \\
  University of York, York, UK \\
  \texttt{charmaine.barker@york.ac.uk} \\\And
  Dimitar Kazakov \\
  University of York, York, UK \\
  \texttt{dimitar.kazakov@york.ac.uk} \\}

\date{}

\begin{document}
\maketitle
\begin{abstract}
The presence of specific linguistic signals particular to a certain sub-group can become highly salient to language models during training. In automated decision-making settings, this may lead to biased outcomes when models rely on cues that correlate with protected characteristics. We investigate whether prompting ChatGPT to rewrite text using simplification, neutralisation, localisation, and formalisation can reduce demographic signals while preserving meaning. Experimental results show a statistically significant drop in location classification accuracy across multiple models after transformation, suggesting reduced reliance on group-specific language. At the same time, sentiment analysis and rating prediction tasks confirm that the core meaning of the reviews remains greatly intact. These results suggest that prompt-based rewriting offers a practical and generalisable approach for mitigating bias in text classification.
\end{abstract}

\section{Introduction}
\label{sec:Introduction}
People from different regions or social groups often express themselves in distinct ways—using certain words, phrases, or stylistic conventions more frequently. These language patterns can become embedded in textual datasets and may be inadvertently picked up by Machine Learning (ML) models during training, especially when those patterns correlate with sensitive attributes like nationality or ethnicity.

In some applications, such as product or movie reviews, these differences might be harmless. However, in high-stakes domains such as automated résumé screening~\citep{nimbekar2019automated}, essay scoring~\citep{ramesh2022automated}, or social media analysis for recruitment~\citep{harirchian2020personality, moraes2020personality}, group-specific language can lead to biased outcomes. If a model learns to associate certain linguistic features with favourable predictions, it may inadvertently encode protected characteristics, raising significant fairness concerns.

Previous work in bias mitigation has focused largely on collecting more representative data, reweighting features, or adjusting model outputs. In contrast, we explore whether standardising the text itself can help. Specifically, we propose rewriting inputs to remove or reduce the linguistic cues that signal the author’s background—before model training even begins.

With the advent of Large Language Models (LLMs) like ChatGPT~\citep{ChatGPT}, it is now possible to perform high-quality text rewriting at scale. In this work, we evaluate whether prompting ChatGPT to revise text can reduce the model’s ability to infer demographic information. If classification accuracy drops after transformation, and the core meaning of the text is preserved, it suggests that text rewriting may be a promising tool for bias mitigation in NLP.

\section{Related Work}
\label{sec:Related Work}
Textual data is prone to bias, especially when language use correlates with demographic attributes. As Machine Learning (ML) systems are deployed in high-stakes domains like hiring~\citep{nimbekar2019automated} and education~\citep{ramesh2022automated}, concerns about fairness have grown.

Several studies have demonstrated how language patterns vary across social groups. For instance,~\citet{jørgensen2015challenges} investigated the impact of African-American Vernacular English (AAVE) on Natural Language Processing (NLP) tools, showing that spelling variations and idiomatic expressions (e.g., \textit{dis}, \textit{dat}, \textit{loc’d out}) degraded the performance of part-of-speech (POS) taggers. Similarly,~\citet{hovy2015tagging} showed that model accuracy varied based on the author's age, highlighting how social group membership can affect NLP outcomes. In both cases, the models implicitly overfit to dominant language patterns, leaving minority dialects underrepresented and poorly handled.

Mitigation strategies often begin at the data collection stage. \citet{bender2018data} argue for careful documentation of dataset composition and context through data statements, which promote awareness of which groups are represented. Post hoc techniques such as oversampling minority groups or downsampling dominant ones can also help balance training data. However, even balanced datasets may include linguistic cues tied to regional or social identity, making it possible to infer protected attributes from writing style. To address this, \citet{deshpande2020mitigating} propose a debiasing technique called \textit{fair tf-idf}, where the importance of each word is reweighted based on its differential prevalence across groups. Words strongly correlated with a specific group (e.g., \textit{India} in Indian resumes) are downweighted, reducing the chance of models latching onto sensitive group markers.

Recent advances in large language models (LLMs), particularly GPT-3.5-turbo as used in ChatGPT~\citep{ChatGPT}, have made controlled rewriting both accessible and effective. Studies have shown that these models outperform state-of-the-art sentence simplification systems across multiple benchmarks, achieving quality ratings comparable to those of human annotators~\citep{feng2023sentence,kew2023bless}. They excel at reducing syntactic complexity, omitting non-essential content, and adjusting tone while preserving the core meaning. Their potential is evident in applications like medical communication: \citet{jeblick2023chatgpt} used ChatGPT to simplify radiology reports, which radiologists confirmed remained accurate and clear.

These capabilities make LLMs well-suited for stylistic rewriting aimed at suppressing group-identifying cues. This work explores text transformation as a means of reducing such cues before model training. By standardising how individuals express similar content, we aim to minimise the presence of linguistic signals that could inadvertently reveal protected attributes. This shifts the focus from balancing data distributions to modifying surface-level expression, offering a complementary pathway for bias mitigation.

Although prior work has also explored rewriting to reduce bias, existing methods differ significantly in their assumptions and complexity. \citet{mireshghallah2021style} use a generative model with latent style disentanglement, requiring attribute-labelled data, tuned encoders, and domain-specific priors. \citet{tokpo2022text} combine masked language modelling with attribute masking, trained via custom classifiers and objectives. In contrast, we examine whether prompting a general-purpose language model can provide a lightweight alternative—one that requires no task-specific training or labelled attributes. To our knowledge, this is the first work to systematically assess the fairness impact of prompt-based rewriting for text classification.

\section{Methodology}
\label{sec:Methodology}

\subsection{Dataset}
\label{sec:Dataset}

We use the Disneyland Reviews dataset~\citep{DisneyData}, an open-source collection of 42,000 user reviews from the platform Tripadvisor\footnote{https://www.tripadvisor.co.uk/}. Each review has the following features:
\begin{itemize}[leftmargin=12pt]
    \item \texttt{Review\_Text}: The raw review text used for transformation and classification.
    \item \texttt{Reviewer\_Location}: The country of origin of the reviewer, used as the protected characteristic of the author, and therefore the target label for our classification task.
    \item \texttt{Rating}: The review's star rating, used later for semantic consistency and retention checks.
\end{itemize}
This dataset is ideal for our study as it contains location-labelled, opinion-rich user-generated text, allowing us to test both bias reduction and semantic fidelity. To reduce computational cost and address class imbalance, we sampled a stratified subset of 10,000 reviews, selecting 2,500 each from the US and UK, 1,000 from Australia and Canada, and 500 from each of six other frequent locations. This preserved geographic diversity while reducing skew. Reviews were truncated to 100 words to standardise outliers and input length. This also ensured comparability with ChatGPT outputs, which were capped at 50 tokens to reflect the brevity typical of simplified or neutralised text, and to reduce stylistic drift.

\subsection{Text Transformation via ChatGPT}
\label{sec:Text Transformation via ChatGPT}
We define six dataset variants, $T_0$ to $T_5$, to test how textual rewriting affects regional inference. $T_0$ contains original reviews, and $T_1$ anonymises them by removing named entities, using Named Entity Recognition (NER). The remaining variants ($T_2$–$T_5$) are rewritten using ChatGPT, following these hypotheses:

\begin{itemize}[leftmargin=12pt]
    \item $H_1$ (Simplification): Using simpler grammar and vocabulary reduces dialectal complexity and regional cues.
    \item $H_2$ (Neutralisation): Removing location-specific terms suppresses regional indicators.
    \item $H_3$ (Localisation): Rewriting all text in a uniform dialect (e.g., British English) reduces inter-group variation.
    \item $H_4$ (Stylistic Shift): Formalising tone removes informal markers tied to dialect, with meaning preserved.
\end{itemize}

For $T_2$ through $T_5$, we use ChatGPT (gpt-3.5-turbo) to generate the transformed versions of each review. Each transformation is applied to the anonymised input ($T_1$) using a fixed prompt.

The specific prompts used for each transformation are as follows:
\begin{itemize}[leftmargin=12pt]
    \item $T_2$: \textit{Simplify this text in 50 tokens or less}
    \item $T_3$: \textit{Remove author nationality cues and use globally neutral English without idioms or slang in 50 tokens or less}
    \item $T_4$: \textit{Make this sound like it was written by an Englishman in 50 tokens or less}
    \item $T_5$: \textit{Rephrase this for a professional audience using formal language in 50 tokens or less}
\end{itemize}

These prompts span a range of linguistic edits, from light simplification to explicit de-biasing, allowing us to assess which styles most reduce demographic signal. Table~\ref{tab:transformations} summarises each variant.

\begin{table*}[htb]
\centering
\resizebox{0.7\textwidth}{!}{%
\begin{tabular}{|c|c|c|}
\hline
Variant & Description & Example Output \\
\hline
$T_0$ & Original text & \textit{The park was awesome and we can not wait to go back.} \\
$T_1$ & Anonymised & \textit{The park was awesome and we can not wait to go back.} \\
$T_2$ & Simple English & \textit{It was awesome. Excited to return.} \\
$T_3$ & Neutral English & \textit{The park was impressive and we look forward to returning.} \\
$T_4$ & British English & \textit{The park was splendid. Can't wait to return.} \\
$T_5$ & Formal English & \textit{We found the park to be excellent and look forward to returning.} \\
\hline
\end{tabular}
}
\caption{Dataset variants along with examples of each used in this study.}
\label{tab:transformations}
\end{table*}
\vspace{-12pt}
\subsection{Location Classification Task}
\label{sec:Location Classification Task}
To measure the extent to which regional cues persist across transformations, we train two binary classifiers to predict the location of a review's author, using only the review text. These models are chosen to reflect both traditional feature-based and modern neural approaches to text classification. The first is an XGBoost model trained on TF-IDF vectors~\citep{chen2016xgboost}. The second is a bidirectional LSTM~\citep{graves2012long} with pre-trained 200-dimensional GloVe embeddings, two stacked LSTM layers (64 and 32 units), and dropout and batch normalisation. A version of each classifier is retrained on each transformed dataset variant $T_i \in \{T_0, T_1, ..., T_5\}$ using early stopping and learning rate scheduling. Classification accuracy serves as an indicator of how much location-specific signal remains detectable in each version. Each classifier is evaluated on a held-out test set across ten stratified 80/20 train/test splits, with mean accuracy and variance reported. A substantial drop in performance from $T_0$ and $T_1$ to the transformed versions ($T_2$–$T_5$) suggests that dialectal or regional features have been effectively suppressed.

\subsection{Semantic Preservation Check}
\label{sec:semantic-preservation}
To assess whether text transformations preserve the underlying meaning of the reviews, we conduct two additional sentiment-based evaluations: (1) a VADER~\citep{VADER} sentiment comparison, which uses the compound sentiment score to evaluate whether the polarity (positive, neutral, or negative) remains consistent between the original and transformed texts; and (2) a rating prediction task, where classifiers are trained to infer the original 1--5 star rating from each transformed version. High agreement in sentiment polarity and prediction error in the rating task suggest that the essential content and tone of the review remain intact.

\section{Results}
\label{sec:Results}
\subsection{Location Classification Results}
\label{sec:Location Classification Results}

\begin{table*}[h]
\centering
\resizebox{0.8\textwidth}{!}{%
\begin{tabular}{|c|c|c|c|c|}
\hline
Variant & XGBoost Accuracy & $\Delta$ XGBoost Accuracy & BiLSTM Accuracy & $\Delta$ BiLSTM Accuracy \\
\hline
$T_0$ (Original) &  $50.45 \pm 0.50$& – &  $44.96 \pm 1.17$& – \\
$T_1$ (Anonymised) &  $47.53 \pm 0.79$&  $-2.92 \pm 0.69$&  $43.91 \pm 0.78$&  $-1.05 \pm 0.76$\\
$T_2$ (Simple English) &  $39.16 \pm 0.49$&  $-11.29 \pm 0.48$&  $38.91 \pm 0.89$&  $-6.04 \pm 1.01$\\
$T_3$ (Neutral English) &  $32.88 \pm 0.84$&  $\bm{-17.57 \pm 1.17}$&  $33.32 \pm 0.90$&  $\bm{-11.63 \pm 1.68}$\\
$T_4$ (British English) &  $35.55 \pm 1.07$&  $-14.91 \pm 1.22$&  $35.52 \pm 0.63$&  $-9.44 \pm 1.26$\\
$T_5$ (Formal English) &  $36.86 \pm 0.88$&  $-13.59 \pm 0.75$&  $36.46 \pm 0.78$&  $-8.50 \pm 1.17$\\
\hline
\end{tabular}
}
\caption{Location classification accuracy (\%) for XGBoost and BiLSTM models across all dataset variants. $\Delta$ columns report the average drop in accuracy relative to $T_0$, where negative $\Delta$ values indicate a performance drop relative to $T_0$. Standard deviations for $\Delta$ reflect variability across runs. The largest performance drop is highlighted in \textbf{bold}.}
\label{tab:core_results}
\end{table*}

The results of the location classification task are presented in Table~\ref{tab:core_results}\footnote{The code and generated dataset can be found at: \url{https://github.com/CharmaineBarker/Mitigating-Bias-in-Text-Classification-via-Prompt-Based-Text-Transformation}}. On the original dataset ($T_0$), classifiers achieve accuracies of $50.45\% \pm 0.50$ (XGBoost) and $44.96\% \pm 1.17$ (BiLSTM), indicating that a substantial amount of regional information is embedded in the raw text—enough for the model to infer the author’s nationality at well above the $25\%$ majority-class baseline. Anonymisation alone ($T_1$) results in a modest performance drop of approximately $-2.92\%$ for XGBoost and $-1.05\%$ for BiLSTM, confirming that direct removal of named entities eliminates some location-specific signals, but not all. However, in $T_2$, larger declines in accuracy were observed: over $11\%$ for XGBoost and $6\%$ for BiLSTM. This suggests that simplification paraphrases or removes indirect cues that models previously relied on.

More targeted prompt design can further reduce the ability of models to infer location. The most effective transformation in this regard is $T_3$ (Neutral English), where classifier accuracy drops to $32.88\%$, a decline of $17.57\%$ relative to the original text's classifier. This prompt explicitly instructed the model to remove author nationality cues, likely leading to more consistent removal of location-indicative language. The second most effective transformation is $T_4$ (British English), which results in a $14.91\%$ drop in accuracy. In this case, although the model introduced distinctly British phrasing, it did so uniformly across all authors, effectively neutralising the dataset by making all reviews appear similarly British. As a result, the classifier was less able to distinguish between authors' locations. For a deeper analysis of the lexical shifts driving these changes, see Section~\ref{sec:Feature Importance Shift Analysis}.

All observed drops in classification accuracy across $T_2$ through $T_5$ were found to be statistically significant using paired $t$-tests ($p < 0.05$) for both the XGBoost and BiLSTM models, confirming that the reductions in predictive performance are unlikely to be due to random variation.

\subsection{Sentiment Preservation Results}
\label{sec:Sentiment Preservation Results}

\begin{figure*}[htbp]
\centering
    {\includegraphics[width=0.9\textwidth]{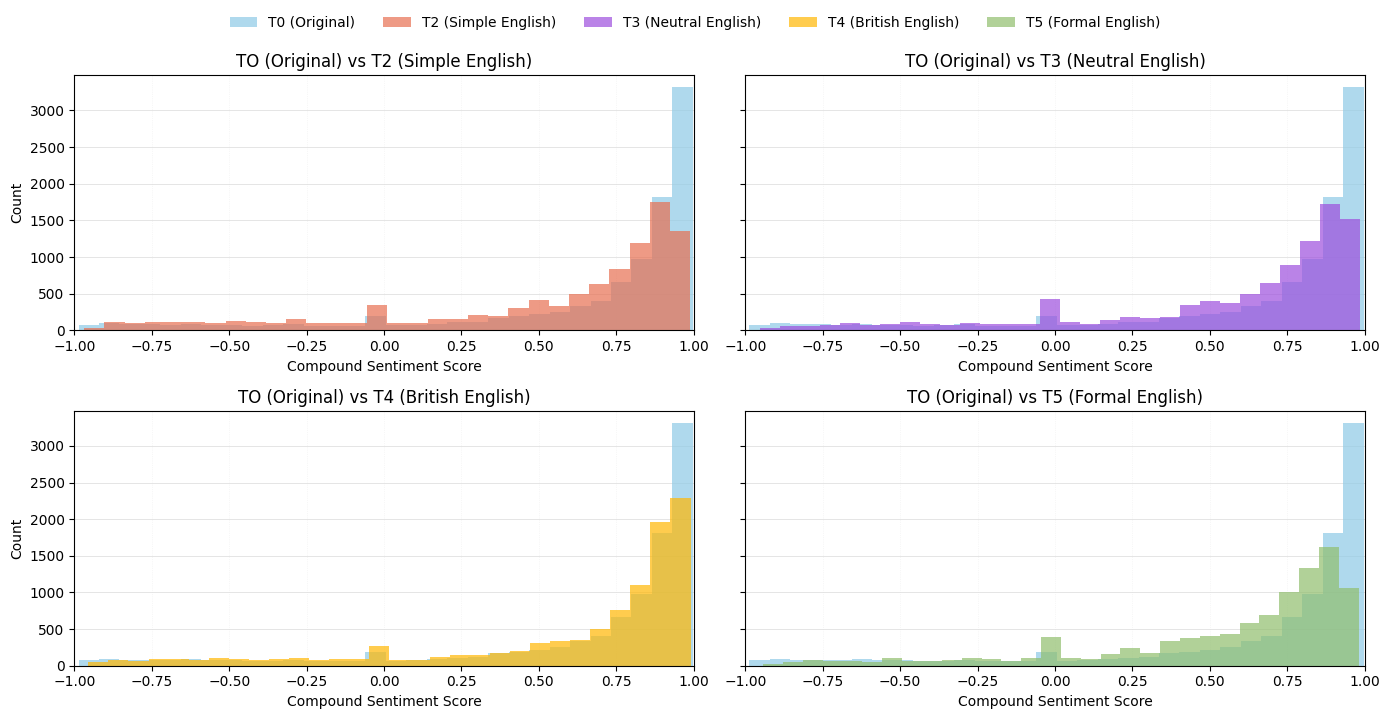}}
    \caption{Sentiment distributions for the original reviews ($T_0$) and four simplified variants ($T_2$–$T_5$), using VADER compound scores. Each subplot compares $T_0$ to a transformed version, with higher histogram overlap indicating better sentiment preservation.}
    \label{fig:sentiment_histogram}
\end{figure*}

Figure~\ref{fig:sentiment_histogram} presents the distribution of VADER compound sentiment scores before and after transformation across the four variants of the original review text ($T_0$). Each subplot overlays the sentiment distribution of $T_0$ (Original) with one of the transformed outputs: $T_2$ (Simple English), $T_3$ (Neutral English), $T_4$ (British English), and $T_5$ (Formal English). Across all transformations, sentiment polarity remains largely consistent, as evidenced by the strong histogram overlap. The $T_1$ (NER-Anonymised) variant is excluded from the figure for visual clarity, but it produced a visually perfect replication of the original sentiment distribution. This outcome is expected, as NER anonymisation modifies only named entities, such as names or locations, which have minimal influence on sentiment polarity.

While all transformations broadly retain the sentiment profile of the original text, $T_3$ (Neutral English) exhibits the largest, though still very modest, divergence. This is likely due to the nature of the prompt, which encouraged the removal of author nationality cues but did not consistently replace them with more neutral options, often resulting in disjointed or incomplete phrasing. $T_4$ (British English) most closely mirrors the sentiment distribution of $T_0$, whereas $T_2$ (Simple English) and $T_5$ (Formal English) tend to be less expressive, possibly due to lexical simplification or more restrained tone.

Despite these subtle differences, the overarching pattern is clear-- all transformed versions closely mirror the sentiment of the original reviews. To further assess semantic similarity, we evaluate how well transformed reviews retain enough information for predicting the 1–5 star rating present in the original dataset. If the core meaning is preserved, then models trained on transformed text should perform comparably to those trained on the original input.

We train both classifiers on each $T_1$–$T_5$, using the rating as the target label. Results in Table~\ref{tab:rating_results} show that rating prediction accuracy never drops after transformation. Some versions, particularly $T_2$ (Simple English) and $T_5$ (Formal English), even exceed baseline accuracy, likely due to the simplification of evaluative content. Notably, the best-performing variants in terms of location being harder to detect are also those that perform the most similar in terms of rating detection to the original. These findings confirm that transformations do not distort the core evaluative meaning of the reviews.

\begin{table*}[h]
\centering
\resizebox{0.8\textwidth}{!}{%
\begin{tabular}{|c|c|c|c|c|}
\hline
Variant & XGBoost Accuracy & $\Delta$ XGBoost Accuracy & BiLSTM Accuracy & $\Delta$ BiLSTM Accuracy \\
\hline
$T_0$ (Original) &  $56.32 \pm 0.71$& – &  $56.05 \pm 0.79$& – \\
$T_1$ (Anonymised) &  $56.43 \pm 0.67$&  $+0.11 \pm 0.86$&  $56.38 \pm 0.79$&  $+0.34 \pm 0.99$\\
$T_2$ (Simple English) &  $57.85 \pm 0.77$&  $+1.54 \pm 0.94$&  $58.76 \pm 0.87$&  $+2.71 \pm 1.06$\\
$T_3$ (Neutral English) &  $56.60 \pm 0.79$&  $+0.28 \pm 0.80$&  $57.59 \pm 0.87$&  $+1.54 \pm 1.13$\\
$T_4$ (British English) &  $56.84 \pm 0.59$&  $+0.52 \pm 0.83$&  $56.88 \pm 0.87$&  $+0.84 \pm 1.23$\\
$T_5$ (Formal English) &  $57.75 \pm 1.24$&  $+1.43 \pm 1.37$&  $57.99 \pm 1.26$&  $+1.94 \pm 1.47$\\
\hline
\end{tabular}
}
\caption{Rating classification accuracy (\%) for XGBoost and BiLSTM models across all dataset variants. $\Delta$ columns report the average difference in accuracy relative to $T_0$, where positive $\Delta$ values indicate a performance increase relative to $T_0$. Standard deviations for $\Delta$ reflect variability across runs.}
\label{tab:rating_results}
\end{table*}

\subsection{Feature Importance Shift Analysis}
\label{sec:Feature Importance Shift Analysis}

\begin{figure*}[htbp]
\centering
    {\includegraphics[width=0.95\textwidth]{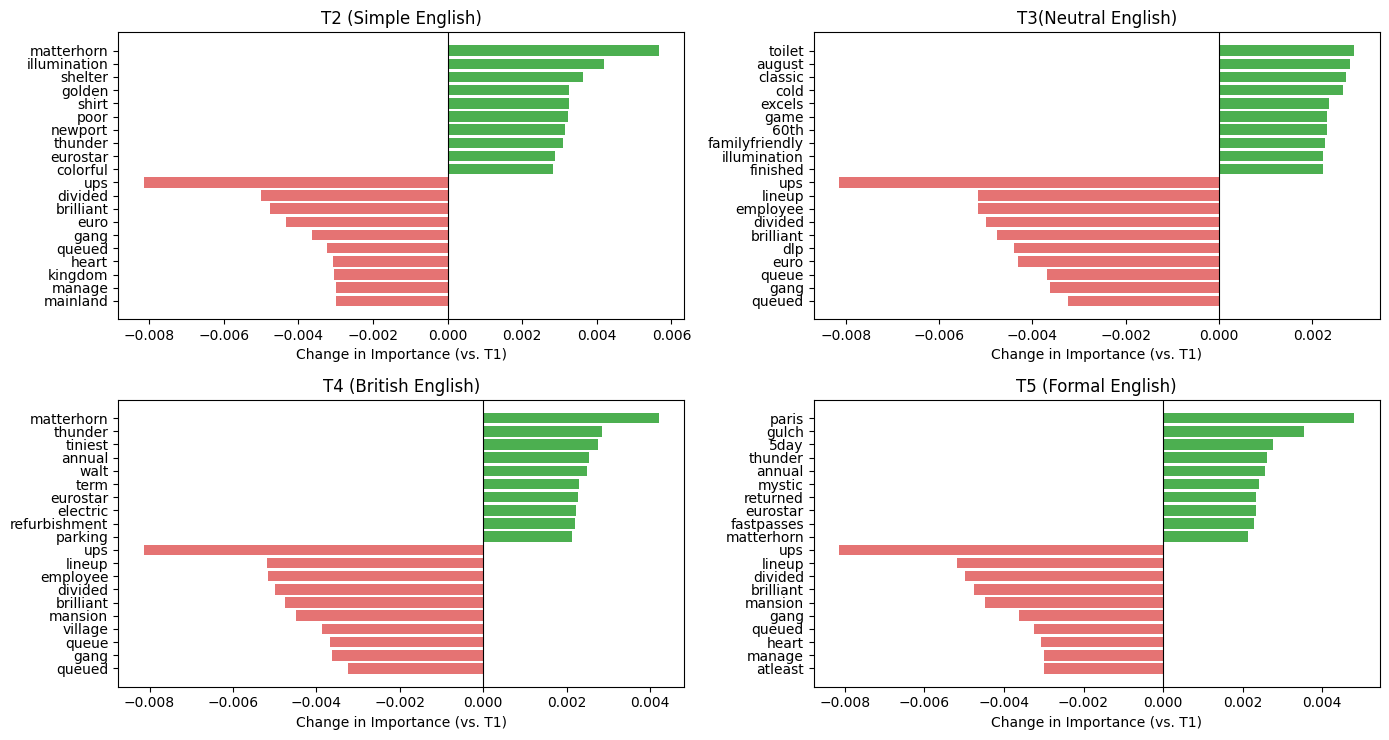}}
    \caption{Change in TF-IDF XGBoost feature importance ($\Delta$ importance) for each transformed version ($T_2$-$T_5$) relative to the anonymised input ($T_1$). Bars show the top 10 word features with the largest increase (green) or decrease (red) in predictive value after transformation.}
    \label{fig:importance_deltas}
\end{figure*}

To better understand \textit{how} ChatGPT's transformations affect classification performance, we examine which words gain or lose predictive value across transformations. Specifically, we focus on changes in their importance to the location classification task. Figure~\ref{fig:importance_deltas} illustrates the top increases and decreases in word importance between the original NER-based input ($T_0$) and each transformed version ($T_2$--$T_5$), based on feature importances derived from XGBoost classifiers trained on each column. Green bars indicate words whose importance increased, while red bars represent those whose importance diminished. 

Complementing the importance scores, we analysed the raw frequencies of the words whose predictive value changed the most after transformation. This allows us to assess whether shifts in model reliance can be explained by reduced usage, complete omission, or replacement of these words in the simplified outputs.

Across all transformations, words such as \textit{queue} and \textit{employee} consistently decline in feature importance. These terms are strongly indicative of specific regional or contextual usage in the original input, and their demotion suggests that the simplification process either removed or neutralised these location-specific cues. For instance, the word \textit{queue} appeared in approximately $\approx13\%$ of the original reviews written by users from the United Kingdom (UK), but in only $\approx1\%$ of those from the United States (US), highlighting its value as a strong dialectal indicator. In the simplified versions, however, its frequency overall presence dropped to just $\approx1\%$ in UK reviews and $\approx0.1\%$ in US reviews, demonstrating a significant loss of regional distinctiveness.

A similar pattern emerges with \textit{employee(s)}, which featured in around $\approx3\%$ of original US reviews but only $\approx0.2\%$ of UK ones. Conversely, UK reviewers used \textit{staff} far more frequently, at a rate of $\approx21\%$ compared to just $\approx6\%$ in US texts. These examples highlight that the words experiencing the greatest drop in classifier importance are often those most directly tied to the author’s linguistic background. Their removal through simplification results in a more lexically neutral version of the text, with fewer cues that a model could exploit to infer author location—thereby contributing to a less biased representation.

Meanwhile, emergent terms such as \textit{matterhorn} and \textit{paris} gained predictive value, suggesting a shift in descriptive focus or content emphasis introduced by ChatGPT’s transformations. For instance, while the word \textit{paris} was entirely removed in the anonymised input ($T_1$), the $T_5$ (Formal English) variant reintroduced contextual cues that implicitly signalled the location, allowing the model to recover some geographical inference. Similarly, \textit{matterhorn} refers to a Disneyland attraction specific to the Anaheim park in California—an entity that NER-based anonymisation was not equipped to detect or redact. Its rising importance likely reflects the diminishing presence of more explicit location markers, prompting the classifier to rely on subtler, domain-specific references that escaped anonymisation. In contrast, $T_3$ (Neutral English) appears to be more effective at eliminating such implicit cues, with these terms occurring far less frequently, resulting in minimal contribution to classifier performance. This suggests that certain transformation styles may offer stronger bias mitigation by actively suppressing geographical indicators.

\section{Conclusion and Future Work}
\label{sec:Conclusion and Future Work}

This paper explored the use of ChatGPT-based text rewriting as an approach to mitigate dialect bias in text. We applied a range of transformation prompts to a regional text dataset and evaluated their impact on classification accuracy, sentiment retention, and lexical shift. The results show that stylistic rewriting can substantially reduce the presence of location-indicative features, leading to more neutral model behaviour. These findings highlight the potential of large language models as tools for bias reduction through controlled rewriting. While promising, our method may face challenges in lower-resource languages or those with rich morphological gender marking, where bias cues are more deeply embedded. Future work will extend this approach to additional prompts and evaluate its effectiveness across broader datasets and linguistic settings.

\bibliographystyle{acl_natbib}
\bibliography{ranlp2023}

\end{document}